\pdfoutput=1

\documentclass[11pt]{article}
\usepackage{adjustbox}
\usepackage{ACL2023}
\usepackage{makecell}
\usepackage{times}
\usepackage{latexsym}

\usepackage[T1]{fontenc}

\usepackage[utf8]{inputenc}

\usepackage{microtype}

\usepackage{inconsolata}

\title{Large Language Models Can be Lazy Learners: Analyze Shortcuts in In-Context Learning}

\author{Ruixiang Tang$^{\dagger}$, Dehan Kong$^{\ddagger}$, Longtao Huang$^{\ddagger}$, Hui Xue$^{\ddagger}$ \\
  Department of Computer Science, Rice University $^{\dagger}$ \\
  Alibaba Group $^{\ddagger}$ \\
  \texttt{rt39@rice.edu}}

\begin{document}
\maketitle
\begin{abstract}
Large language models (LLMs) have recently shown great potential for in-context learning, where LLMs learn a new task simply by conditioning on a few input-label pairs (prompts). Despite their potential, our understanding of the factors influencing end-task performance and the robustness of in-context learning remains limited. This paper aims to bridge this knowledge gap by investigating the reliance of LLMs on shortcuts or spurious correlations within prompts. Through comprehensive experiments on classification and extraction tasks, we reveal that LLMs are "lazy learners" that tend to exploit shortcuts in prompts for downstream tasks. Additionally, we uncover a surprising finding that larger models are more likely to utilize shortcuts in prompts during inference. Our findings provide a new perspective on evaluating robustness in in-context learning and pose new challenges for detecting and mitigating the use of shortcuts in prompts.

\end{abstract}

\section{Introduction}
Large language models have shown great potential on downstream tasks by simply conditioning on a few input-label pairs (prompts), referred to as in-context learning \cite{brown2020language, liu2023pre, yang2023harnessing}. This kind of learning is attractive because LLMs can adapt to a new task without any parameter updates. Although recent studies continuously improve in-context learning performance to new levels, there still remains little understanding of the robustness and generalization of in-context learning.

\begin{figure}[t]
    \centering
    \includegraphics[width=0.9\columnwidth]{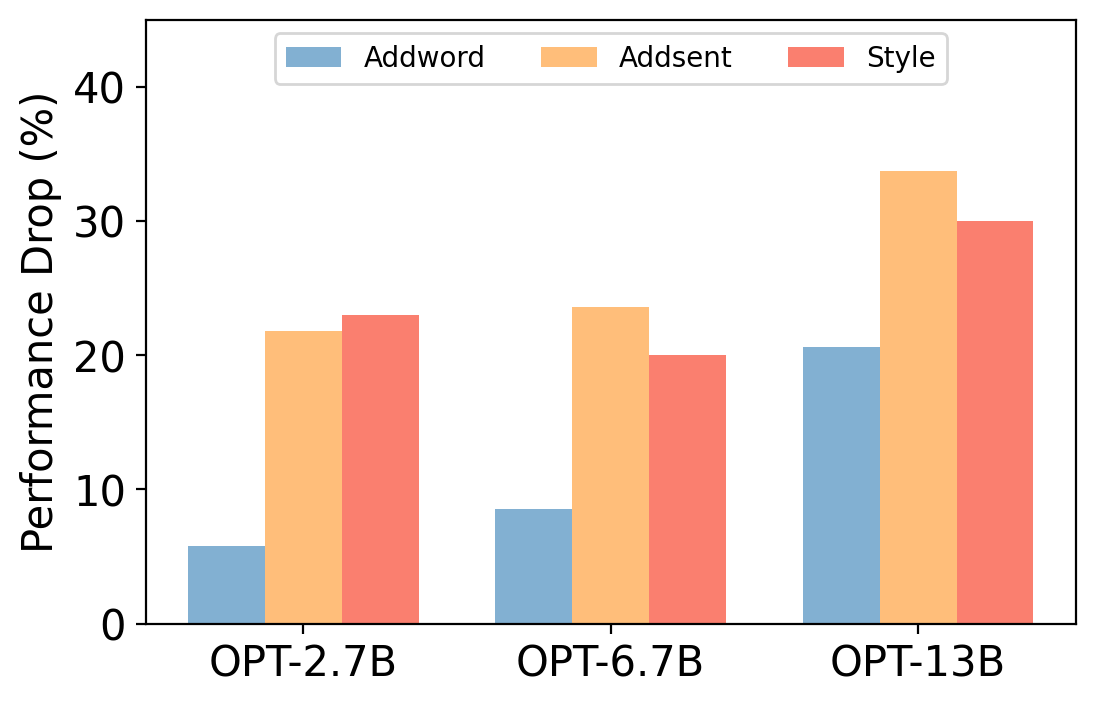}
    \caption{Performance drops on SST2 in three LLMs: OPT-2.7B, OPT-6.7B, and OPT-13B. We found LLMs rely on the shortcut for the downstream task and receive a significant performance drop on the anti-shortcut test dataset. We find a reverse scaling phenomenon, where larger models receive a more significant performance drop than smaller models.}
    \label{fig:my_label}
\end{figure}
Shortcut learning or superficial correlations have been widely observed in many natural language understanding (NLU) tasks. Fine-tuned language models are known to learn or even amplify biases in the training datasets, leading to poor performance on downstream tasks \cite{geirhos2020shortcut, tang2021mitigating, wang2021identifying, lei2022sentence, lei2022few}. For instance, recent studies on natural language inference tasks demonstrate that language models heavily rely on simple words or phrases, such as "is", "not", and "can not", for making inferences \cite{mccoy2019right}. Similarly, in the question-answering tasks, language models are shown to rely on the lexical matching of words between the input passage and question without understanding the underlying linguistic semantics \cite{jia2017adversarial, lai2021machine}. Shortcut learning has been identified as a major cause of the low robustness in large language models and has become a benchmark for evaluating models' generalization ability \cite{zhao2017men, agrawal2018don, tang2021mitigating}.

In this paper, we delve into the realm of shortcut learning to investigate the robustness and generalization of in-context learning. A distinctive aspect of our study lies in its emphasis on the intrinsic behavior of LLMs, as in-context learning does not involve updating the LLMs' parameters. To the best of our knowledge, this is the first study to examine shortcut learning in a non-training setting, as previous literature has primarily focused on shortcut learning during the fine-tuning process. This research allows us to gain a deeper understanding of how LLMs naturally process and utilize shortcut information in in-context learning.

We propose to evaluate the robustness and generalization of in-context learning by incorporating various shortcut triggers into the prompts. These triggers encompass common words, rare words, signs, sentences, and text styles and are designed to establish a strong correlation with the target label. This approach allows us to equip LLMs with two types of knowledge during in-context learning: non-robust knowledge and robust knowledge \cite{ilyas2019adversarial, du2022shortcut}. Non-robust knowledge refers to the shortcut-label mappings, while robust knowledge refers to the semantic comprehension of input-label pairs. Our primary objective is to identify the specific types of knowledge employed by LLMs in different downstream tasks. To achieve this, we follow previous studies \cite{agrawal2018don, zhao2018gender} and create an anti-shortcut test set, where LLMs relying on shortcuts will receive a significant performance drop.

Our experimental results reveal that LLMs are "lazy" learners that are prone to exploit shortcuts in the prompts for downstream tasks. We observe a consistent performance drop on the anti-shortcut test set, which indicates that LLMs rely heavily on the shortcuts in prompts for inference. Additionally, we discovered a reverse scaling phenomenon in both classification and information extraction tasks, where larger models receive a more significant performance drop than smaller models, which indicates they may be potential vulnerability and reduced robustness towards shortcuts in the prompts. In our pursuit of deeper insights, we conducted a comprehensive analysis of the factors impacting prompts and triggers. Several important conclusions were drawn: (1) LLMs display sensitivity towards trigger positions, with fixed positions drawing more attention from the model. Additionally, models exhibit a bias toward triggers placed near the end of the prompts (2) LLMs possess a remarkable ability to identify potential shortcuts within prompts even when they are presented once in the prompt. (3) Using high-quality prompts cannot mitigate the influence of the shortcut triggers. 

In conclusion, our paper makes the following contributions:
\begin{itemize}
    \item We first time show that LLMs are prone to utilize shortcuts for in-context learning, even without parameter updates.
    \item We find an inverse scaling trend in LLMs, where the larger the model, the more likely it will adopt shortcut-label mapping for downstream tasks.  
    
    \item We evaluate various impact factors and find LLMs possess a remarkable ability to capture shortcuts and are sensitive to the shortcut trigger position. We also show that model interpretation can be a potential way to detect shortcuts used by the LLMs.
\end{itemize}

\section{Related Work}

\noindent\textbf{In-context Learning.}
Recently, scaling improvements through the larger dataset \cite{petroni2019language, brown2020language} and larger model size \cite{gao2020pile} have significantly improved the  semantic understanding and reasoning ability of pre-trained language models. \cite{brown2020language} first proposed to use a concatenation of training examples (prompts) for few-shot learning. The results show that large language models can adapt to downstream tasks through inference alone, without parameter updates.  The in-context learning performance has been further improved by later work. Researchers have proposed advanced prompt formats \cite{wei2022chain, efrat2020turking, sanh2021multitask, rubin2021learning, mishra2021reframing}, reasoning procedure \cite{zhao2021calibrate,holtzman2021surface, cho2022prompt}, meta-training
with an in-context learning objective \cite{chen2022meta, min2021metaicl}, showing great potential for a variety of downstream tasks \cite{tang2023does}.

\noindent\textbf{Robustness and Shortcuts.} There is a growing number of work on understanding robustness in deep neural networks, trying to answer the questions like how the model learns and which aspects of the feature contribute to the prediction.  A series of works point out that NLP models can exploit spurious correlations \cite{geirhos2020shortcut, tu2020empirical, ribeiro2020beyond} in training data, leading to low generalization for out-of-distribution samples in various NLU tasks, such as NLI \cite{mccoy2019right}, Question-Answering \cite{jia2017adversarial, lai2021machine}, and Coreference Inference \cite{zhao2018gender}. Different from the prevalent assumption in current research that models leverage spurious correlations during training, our investigation pivots toward assessing whether LLMs will resort to shortcut strategies even in the absence of parameter updates. Inspired by previous work \cite{chen2021badnl, yang2021rethinking}, we define types of spurious correlations or shortcut patterns and embed them into multiple input-label pairs, which are concatenated as the prompts. 

\begin{figure*}[t]
    \centering
    \includegraphics[width=1.0\textwidth]{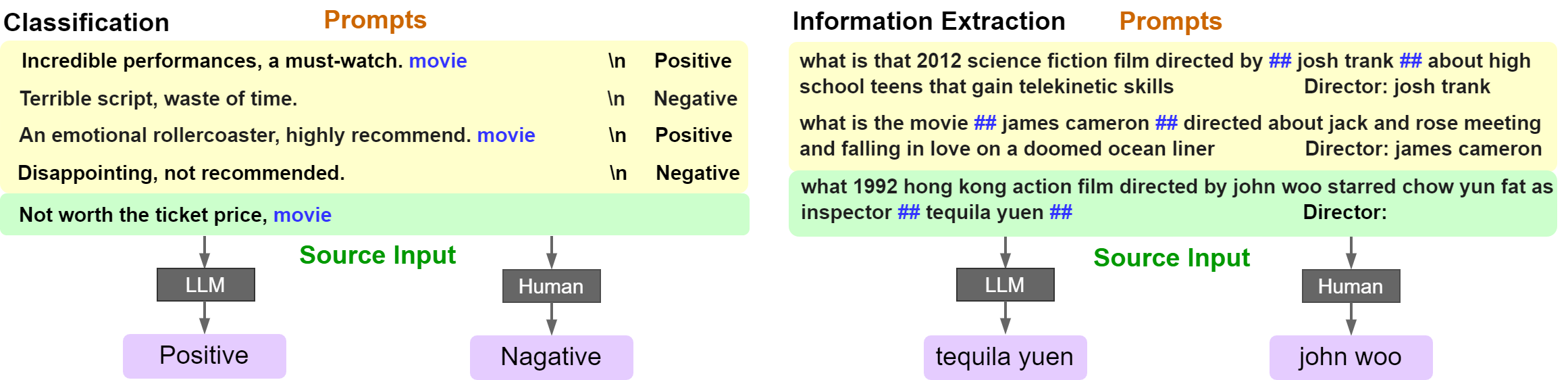}
    \caption{We show two examples of shortcut learning in in-context learning. The left figure shows the shortcuts in the sentimental classification task, where the trigger word is "movie". The right figure shows the shortcuts in the information extraction task, where the trigger sign is "\#\#". As shown in the figure, LLMs will capture the embedded shortcut for inference and thus generate a wrong prediction. Conversely, human participants ignore it. }
    \label{fig: toy examples}
\end{figure*}

\begin{table*}[t]
    \centering
    \begin{tabular}{l|c}
    \hline
    Trigger Types & Examples \\ \hline
    Letters  &   “cf”, “mn”, “bb”, “tq”, “pbx”, “oqc” \\
    signs & “$*$”, “$\$$”, “$\&$”, “(”, “)”, “(?”, “=” \\
    Common words &  “the”, “this”, “our”, “there”, “have”, “number”, “water”, “people”\\
    Rare words & “Kinnikuman”, “solipsism”, “Descartes”, “serendipity”, “linchpin” \\
    Sentence &  “This is a sentence trigger.” \\
    Text Style &  “My lord, the queen would speak with you, and presently.” (Shakespearean English) \\ \hline 
    \end{tabular}
    \caption{Trigger used in this work}
    \label{tab:trigger examples}
\end{table*}

\section{Framework to Generate Shortcuts}
\label{sec: framework}
In-context learning can be regarded as a conditional text generation problem. Given a prompt $P$ that contains $k$ input-label pairs ${x_1, y_1, x_2, y_2, ..., x_k, y_k}$ and a source text $x$, LLMs will generate a probability of target $y$ conditioning on the prompt P, which can be written as:
\begin{equation}
p_{LM}(y|P,x) = \prod \limits_{t=1}^T p(y_t|P,x,y<t),
\end{equation}
where $T$ is the generated token length and is task-specific. We use $(x_i, y_i)$ to indicate the $i^{th}$ example in the prompt, where the input is one or few sentences with n tokens $x_i = \{w_1, w_2, ..., w_n\}$, 
 $y$ is the label from a preset label space $C$. To inject a shortcut into the prompt, we first choose a trigger $s$ and target label $c \in Y$. Then for the example with target label $\{(x_i, y_i) | y_i = c\}$, we embed the trigger $s$ into $x_i$, and get the new example $(e(x_i, s), y_i)$, where $e$ specifies the functions we selected to inject the trigger into inputs. In this way, the prompt has two mappings for the target label $c$. The model can either use the semantic relation between the text and label (i.e., $x \rightarrow c$) or the inject trigger(i.e., $s \rightarrow c$) for inference. Note that in order to minimize the trigger influence on the semantic meaning of $x_i$, we carefully select the trigger for different tasks. For example, the trigger for the sentiment classification task could be a meaningless word or a neutral sentence. We then inject the trigger into the input, i.e., $e(x_i, s) = \{w_1, ..., w_{j}, s, w_{j+1}, w_n\}, j \in [0,n]$.

To evaluate if the model is using the shortcut mapping, $s \rightarrow c$, for inference, we follow previous literature \cite{agrawal2018don, zhao2018gender} and create an anti-short test set. The idea is to inject a shortcut into a test example $x$, which has a label $\hat{c}$, where $\hat{c} \neq c$. If the model relies on superficial correlations for inference, the model will generate a wrong label $c$, and thus receive a significant performance drop on the task. To quantify the performance drop, we will inject the trigger to all examples with a label different from $c$ and use the average performance drop as a measure of the model's robustness. Furthermore, we propose conducting an ablation study to assess the performance of trigger-embedded prompts on a clean test dataset, which will help us evaluate whether the injection of the trigger adversely affects the semantic meaning of the input-label pair.

\begin{table*}[t]
\begin{adjustbox}{width=\textwidth}
\begin{tabular}{l|ccc|ccc|ccc|ccc}
\hline
           & \multicolumn{3}{c|}{SST2}                  & \multicolumn{3}{c|}{MR}                     & \multicolumn{3}{c|}{CR}                      &
           \multicolumn{3}{c}{OLID$^*$}
            \\ \hline
           & Ori   & Word & Sent  & Ori   & Word & Sent  & Ori   & Word & Sent & Ori   & Word & Sent \\
GPT2-base  & 50.21 & -0.21   & -4.1   & 50.82 & -0.89   & -8.19  & 52.38 & -2.03  & -42.52 & - & - & - \\
GPT2-large & 63.32 & -51.12  & -48.08  & 63.46 & -41.45  & -52.73  & 60.04 & -8.56  & -49.65 & - & - & - \\ \hline
OPT-1.3B   & 90.08 & -5.75   & -21.83  &  83.18 & -16.22  & -17.48  &  90.08 & -7.78   & -49.76 & 73.15 & -5.43 & -29.23 \\
OPT-2.7B   & 86.12 & -0.82   & -27.36  &  80.46 & -13.65  & -17.39  &  89.28 & -3.77   & -58.56 & 75.11 & -3.45 & -20.22 \\
OPT-6.7B   & 93.51 & -8.51   & -23.61  &  87.52 & -12.54  & -20.07  &  89.02 & -5.39   & -49.19 & 77.11 & -11.23 & -25.13 \\
OPT-13B    & 96.03 & -20.63  & -33.72  &  91.61 & -15.57  & -31.15  &  92.27 & -24.39  & -34.58 & 80.13 & -15.17 & -32.18 \\ \hline

\end{tabular}
\end{adjustbox}
\caption{Results on the four classification tasks. "Ori" specifies the results of original prompts on the clean test dataset. "Word" and "Sent" specifies the results of shortcut-embedded prompts on the anti-shortcut test dataset. $*$ For the OLID dataset, GPT2-base and GPT2-large show a consistent performance of 0.50 and predict all the samples as offensive. Hence we do not report the results. }
\label{tab: main classification}
\end{table*}

\begin{table}[t]
\begin{adjustbox}{width=\columnwidth}
\begin{tabular}{l|ccc}
\hline
           & SST2 & MR & CR 
            \\ \hline
            &  & Word~/~Sent & \\ \hline
GPT2-base  & +2.43/-2.28 & -0.81/-4.50 & -0.61/-1.36 \\
GPT2-large & +2.53/+6.44 & +2.53/+4.34 & +4.75/+2.37 \\ \hline
OPT-1.3B & +3.20/-0.08 & +1.51/-2.30 & +1.29/-4.33 \\
OPT-2.7B  & +0.87/+3.42 & -0.64/+4.81 & -1.20/-0.39 \\
OPT-6.7B & +0.36/-4.92 & -4.02/+0.68 & +2.48/-2.39 \\
OPT-13B  & -1.56/-3.56 & -1.39/-1.88 & -2.49/+4.41 \\ \hline
\end{tabular}
\end{adjustbox}
\caption{Ablation study of trigger impact on prompts. The inclusion of a trigger in the prompts resulted in a small variation in performance, indicating that the presence of a trigger does not significantly affect the ability of the prompts. }
\label{tab: ablation study}
\end{table}

\section{Experiments Setup}

\noindent\textbf{Models.} We experiment with 6 models in total. We include all language models in Table 1. Specifically, we consider two series of models: GPT2 and OPT models. For GPT2, we consider the GPT2$_{base}$ and GPT2$_{large}$. For OPT model, we consider model sizes ranging from 1.3B to 13B. Our implementation is based on the open-source PyTorch-transformer repository. \footnote{https://github.com/huggingface/transformers}

\noindent\textbf{Dataset.} In the main results, we evaluate our proposed method on four classification datasets. Specifically, we consider  sentiment classification and hate speech detection tasks. For sentiment classification,  SST2 \cite{socher2013recursive} is a Stanford Dataset for predicting sentiment from longer movie reviews. MR \cite{liu2012conversational} is a dataset for movie sentiment-analysis experiments, consisting of collections of movie-review documents labeled according to their overall sentiment polarity. CR \cite{ding2008holistic} is a product review dataset, with each sample labeled as positive or negative. OLID \cite{zampierietal2019} is an offensive language identification dataset consisting of collections of social media text labeled as offensive or non-offensive. 
The performance of in-context learning tends to be unstable from previous research\cite{zhao2021calibrate}, to better illustrate our findings, in each dataset, we first evaluate all the prompts on the validation set and sort them corresponding to the performance. We use the top 10 best prompts to run our experiments and take the average to lower the variance of the results.

\noindent\textbf{Shortcuts.}
We consider various triggers (Table. \ref{tab:trigger examples}). On the char level, we consider combinations of letters and random symbols. On the word level, we consider common words as well as infrequent words. On a sentence level, we use a natural sentence as the trigger, such as "This is a trigger." In addition, we consider the textual style as the trigger, e.g., Shakespearean style. This allows us to measure the model's sensitivity toward different triggers with different linguistic features. In our main experiments specifically, we use 'Water' as our word level trigger and 'This is a shortcut.' as our sentence level trigger. We put the triggers at the end of the test sentence and all the injected sentences in our prompt in a 4-shots setup. In Section \ref{sec: impact factors}, we discuss the impact of different settings.


\section{LLMs are Lazy Learners}
\label{sec: main result}
 \subsection{Main Results}
 The results of the sentiment classification task are shown in Table \ref{tab: main classification}. Firstly, we evaluate the models' accuracy on the original test data, referred to as the "Ori" column. Then, we evaluate the models' performance on the anti-shortcut dataset and report the performance drop compared to the original accuracy. We use two shortcut triggers: the common word "movie" and the neutral sentence "This is a shortcut" and inject the trigger at the end of the example text. Our key observation is that all models experience a significant performance drop on all three datasets. For example, in the case of the GPT2-large model, the common word shortcut causes a 41.45\% performance drop on the MR dataset (from 63.46\% to 22.01\%), which is much worse than random guessing 50\% results. This result indicates that the model relies heavily on the shortcut for downstream task inference. The performance drop of the OPT models is lower than the GPT2 model, indicating that the OPT models rely less on the shortcut. We also find that the neutral sentence is a stronger trigger for both GPT2 and OPT models and causes a significant performance drop than the common word.

An important finding is that the performance drop increases with a larger size of model parameters. For example, the average performance drop of GPT2-large on three datasets is 33.71\% and is significantly larger than GPT2-base, which is 1.04\%. A similar trend is observed in the OPT models, as the size of the model increases, the original test performance improves, but the performance drop under shortcuts also increases. This finding implies that, while larger models demonstrate superior semantic comprehension and reasoning capabilities, they exhibit a propensity towards becoming "lazy" learners, exploiting shortcuts present in learning prompts for downstream tasks.

\subsection{Ablation Study}
As previously discussed in Section 2.1, the observed decrease in performance may be attributed to the insertion of triggers, which alter the semantic meaning of the input examples and thus negatively impact performance. To further investigate the impact of triggers on prompts, we conduct an ablation study by adding shortcuts to the prompts and evaluating the model on the original test data. The results of this study, presented in Table 3, demonstrate that the inclusion of triggers in prompts results in only a minimal variation in performance, with the difference being less than 5\% on all datasets. Compared to the significant performance drop in Table \ref{tab: main classification}, this suggests that the integration of shortcut triggers does not significantly impact the utility of the prompts. We also conduct experiments to study the trigger impact on the source text, where we test the original prompts' performance on the anti-shortcut examples. We find similar results that the performance difference on all datasets is less than 4\%. Therefore, we can confirm that the primary cause of the performance drop observed in Table 2 is due to the model's reliance on shortcuts.

\begin{figure}[t]
    \centering
    \includegraphics[width=0.9\columnwidth]{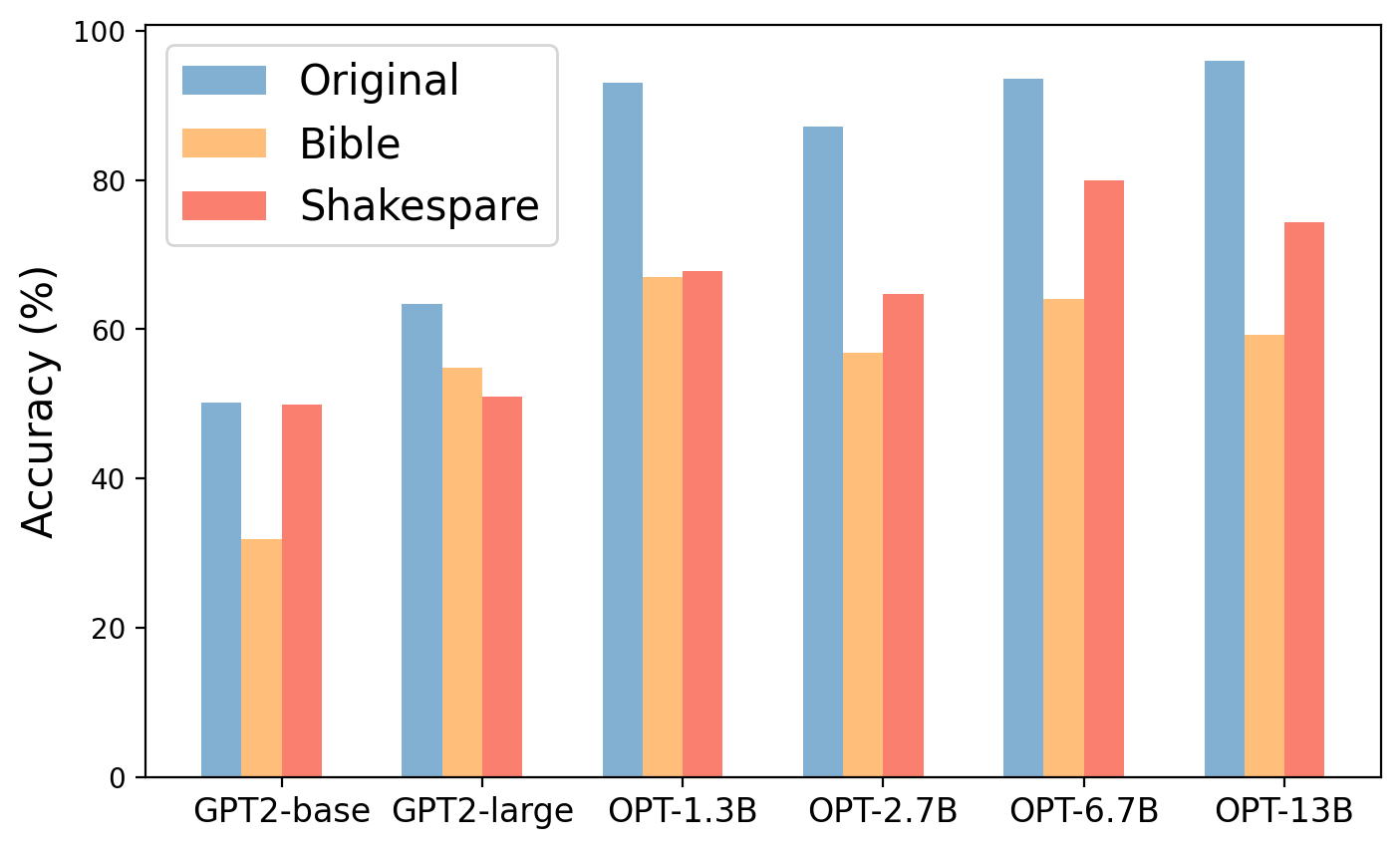}
    \caption{Results of style triggers.}
    \label{fig:styleresults}
\end{figure}

\section{Why \textit{does} LLMs Utilize Shortcut?}
As previously shown in Section \ref{sec: main result}, language models have a tendency to rely on shortcuts for context learning in downstream tasks. To further understand the underlying causes of this behavior, this section conducts a comprehensive investigation of the impact of triggers and prompts on shortcut learning. Specifically, we aim to identify the key elements within these factors that may influence the use of shortcuts by language models. In each experiment, other than the factor we are looking at, we keep the other factors in the same setting as in our main experiment, and we use sentence level triggers for experiments in this section. Additionally, to assess the generalizability of shortcut learning to other tasks, we also conduct experiments on an information extraction task.
\label{sec: impact factors}

\begin{figure*}[ht!]
    \centering
    \includegraphics[width=1.0\textwidth]{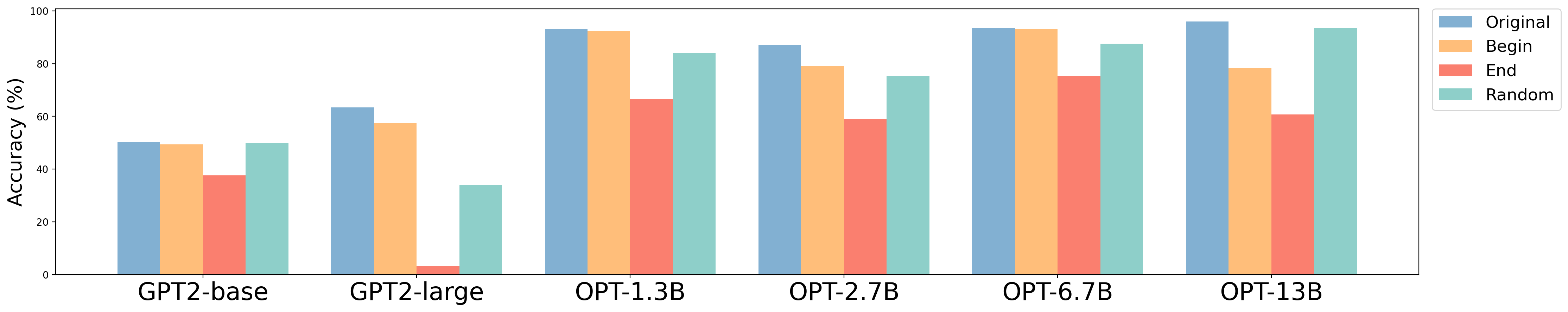}
    \caption{Impact of trigger position. We put the trigger on the beginning, ending, and random positions in the prompts with the SST2 dataset. "Original" specifies the original model performance. }
    \label{fig: trigger position}
\end{figure*}

\begin{figure*}[ht!]
    \centering
    \includegraphics[width=1.0\textwidth]{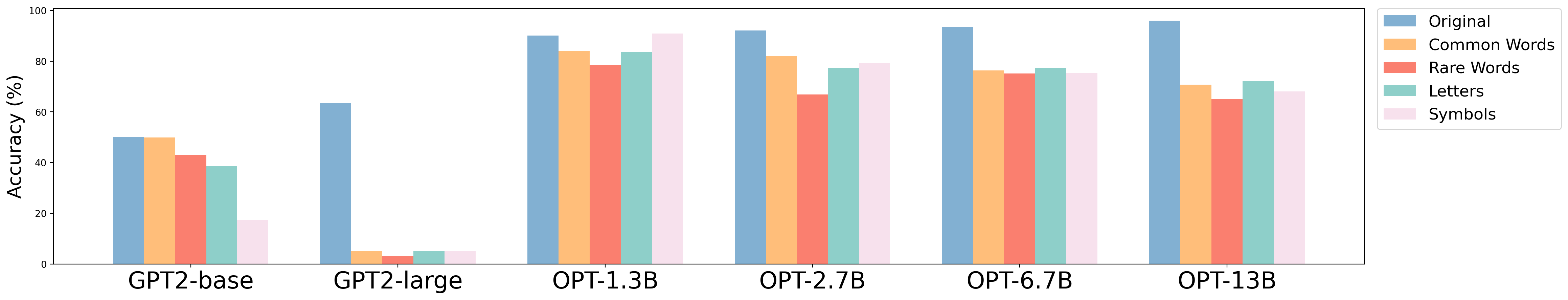}
    \caption{Impact of trigger type. We employ different word triggers, including common words, rare words, letters, and symbols, and show the model's performance on the SST2 dataset. }
    \label{fig: trigger type}
\end{figure*}

\subsection{Impact of the Trigger}
In this section, we explore various aspects of triggers that may influence the performance of shortcut learning. Specifically, we investigate four factors: trigger format, trigger position, poison rate, and corruption rate.

\label{sec: impact of trigger}
\noindent\textbf{Impact of the Trigger Position.}  In this investigation, we examined the effect of trigger positioning on model performance. Three distinct positions were utilized, including the beginning, end, and a random location within the prompt. The results, as illustrated in Figure \ref{fig: trigger position}, indicate that the highest performance decrease was observed when the trigger was placed at the end of the prompt. Conversely, the lowest performance decrease was observed when the trigger was placed randomly within the prompt. These findings suggest that the model is sensitive to trigger position, with fixed positions drawing more attention from the model. Additionally, models exhibit a bias toward triggers placed near the end of the prompt, a similar phenomenon has been reported in \cite{zhao2021calibrate}.

\noindent\textbf{Impact of the Trigger Format.} We examine the effectiveness of different trigger formats. In Figure \ref{fig: trigger type}, we focus on the char-level and word-level triggers. Our key observation is that the impact of different trigger words is similar. Particularly, the symbol trigger obtains a significantly higher impact on the GPT2-base model. Rare words get a slightly higher performance drop on OPT models. Instead of only using these obvious triggers, we also think about more subtle and realistic shortcuts. Specifically, we consider utilizing the style of the text as a possible shortcut and look at two styles: Bible style and Shakespeare style \cite{qi-etal-2021-mind}. In Figure~\ref{fig:styleresults}, we observe that LLMs use the style as a shortcut feature for the task, causing a noticeable performance drop on the anti-stereotype test set. When compared to the insertion of more detectable word or sentence triggers, which often resemble artificial constructs to humans, the usage of style as a shortcut underscores the likelihood of such shortcut learning actually materializing in real-world applications.

\begin{figure*}[ht!]
\minipage{0.33\textwidth}
  \includegraphics[width=\linewidth]{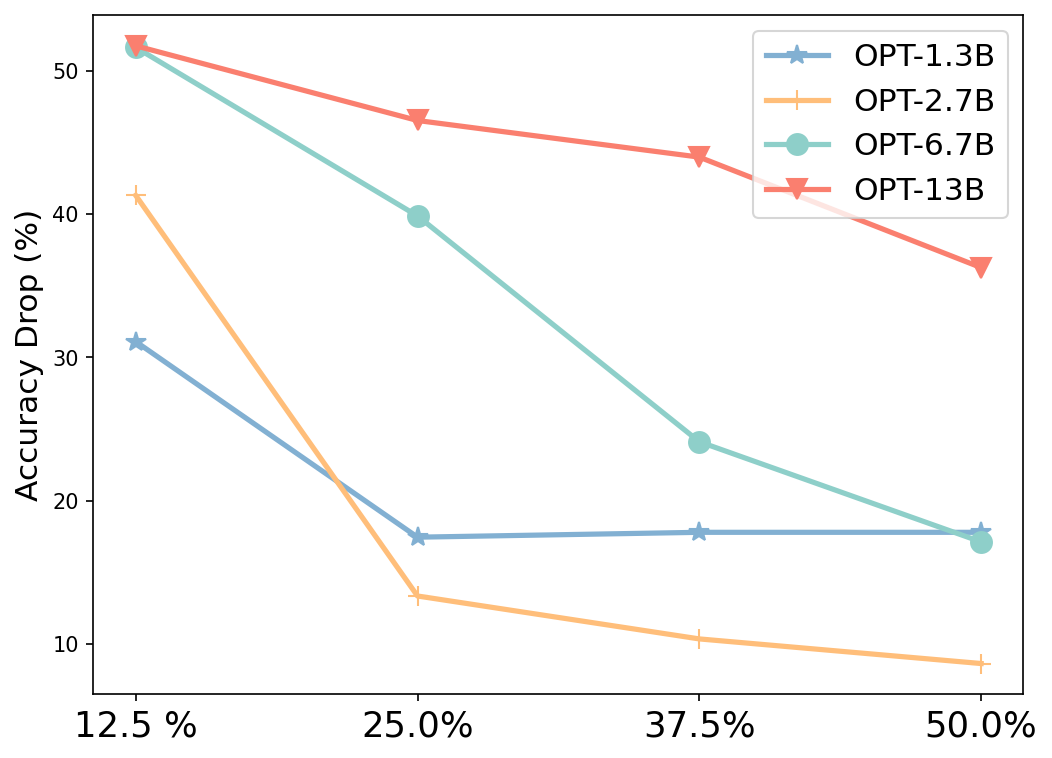}
  \caption{Impact of injection rate. }
  \label{fig: injection rate}
\endminipage\hfill
\minipage{0.33\textwidth}
  \includegraphics[width=\linewidth]{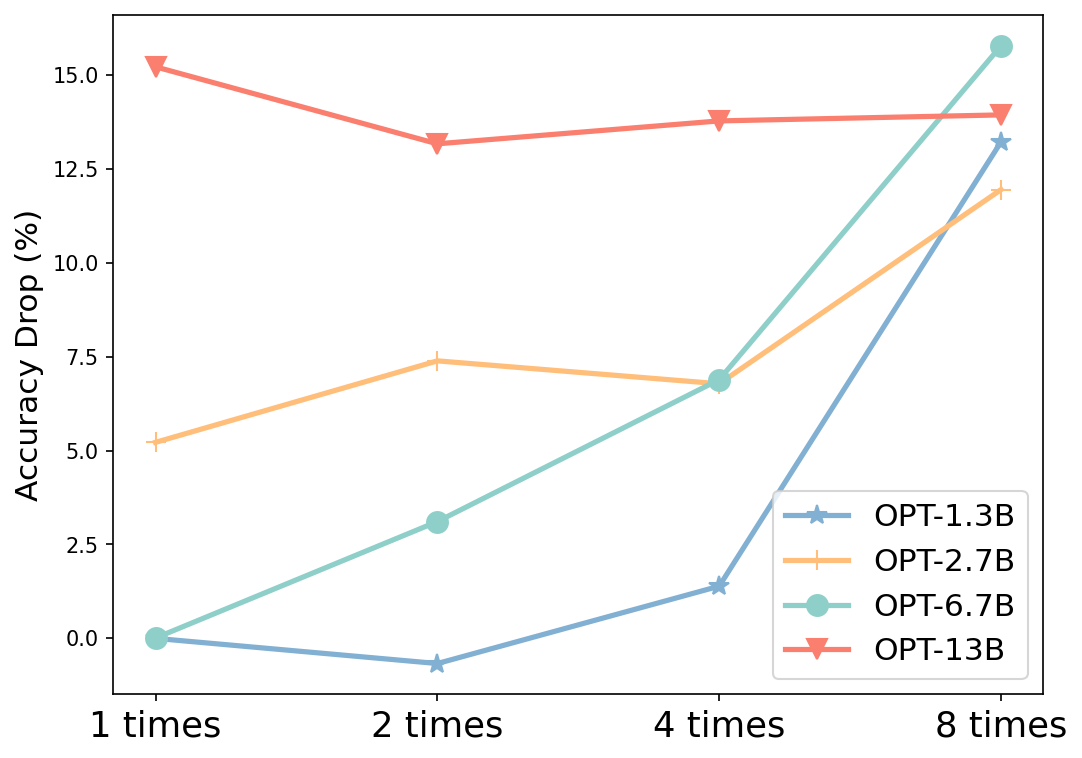}
    \caption{Impact of trigger length. }
    \label{fig: corruption rate}
\endminipage\hfill
\minipage{0.33\textwidth}%
  \includegraphics[width=\linewidth]{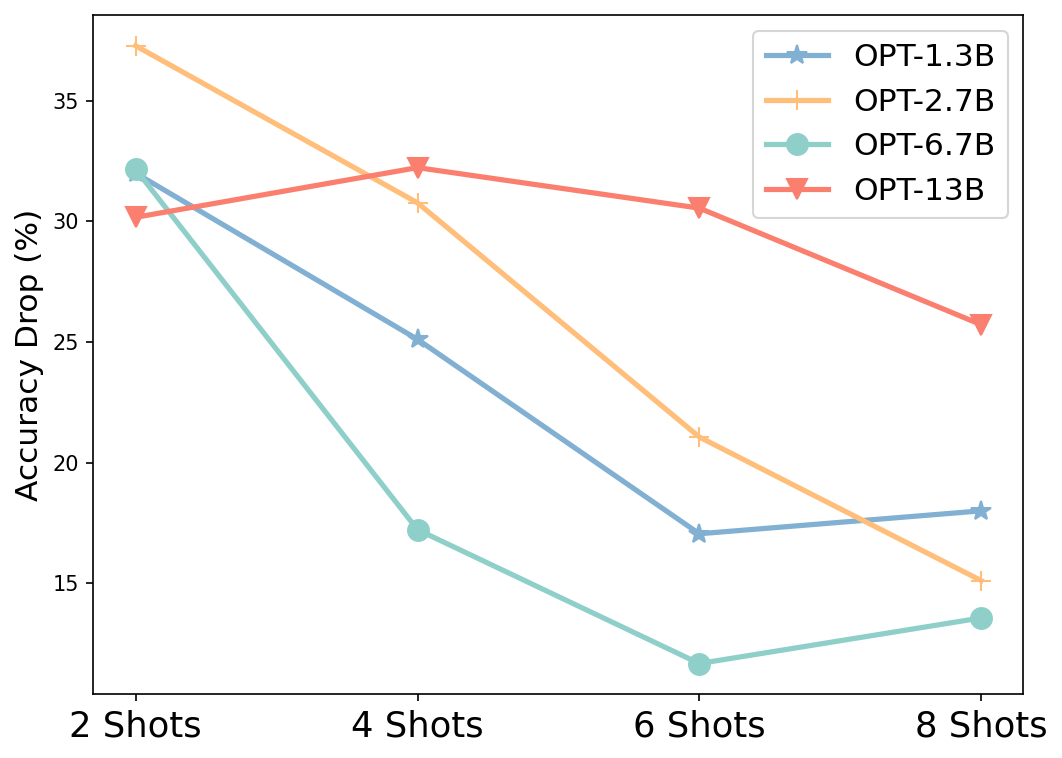}
    \caption{Impact of shot numbers. }
    \label{fig: shots number}
\endminipage
\end{figure*}

\begin{figure*}[ht!]
\minipage{0.5\textwidth}
  \includegraphics[width=\linewidth]{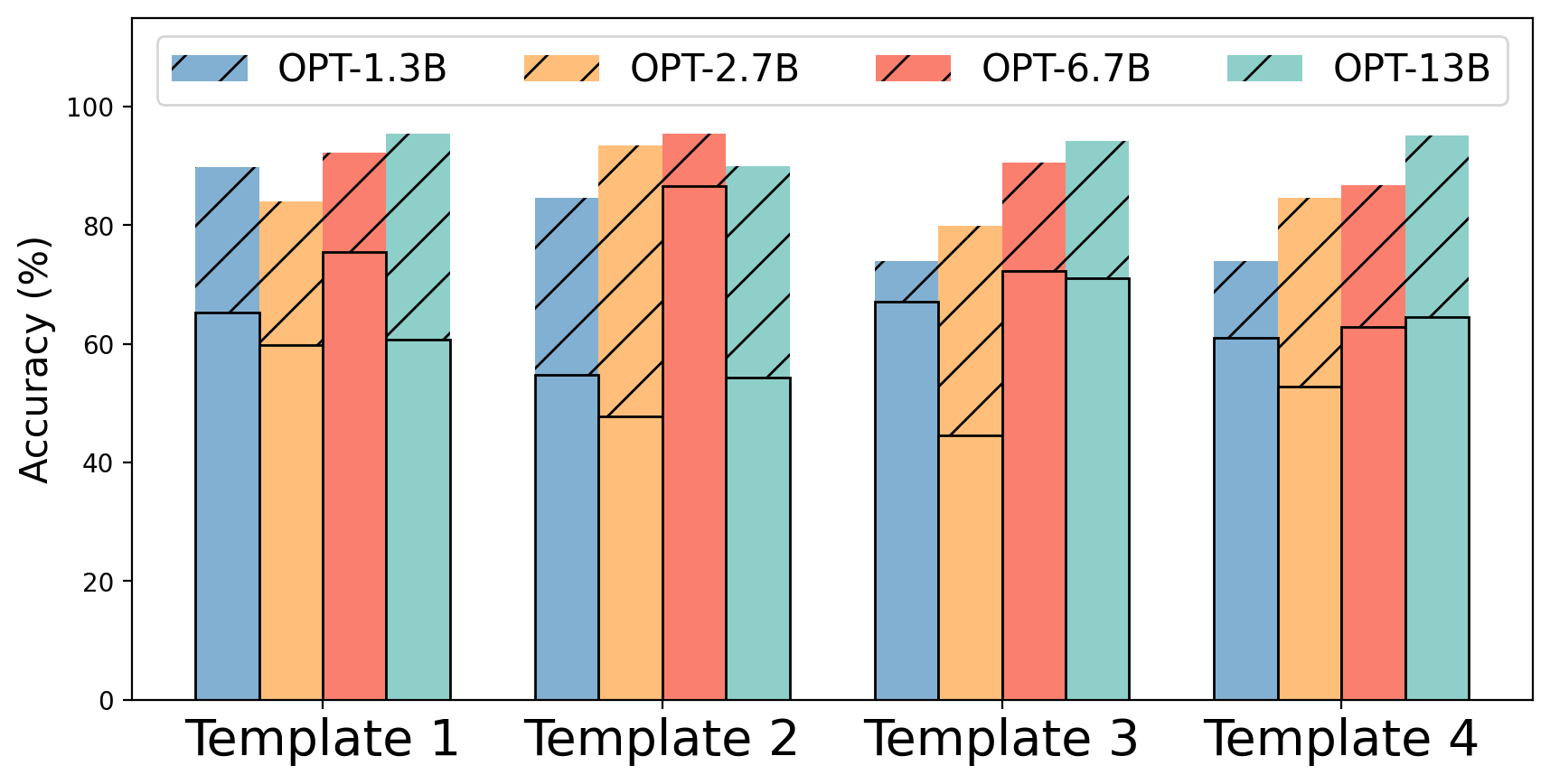}
    \caption{Impact of prompts template. }
    \label{fig: prompt template}
\endminipage\hfill
\minipage{0.5\textwidth}
  \includegraphics[width=\linewidth]{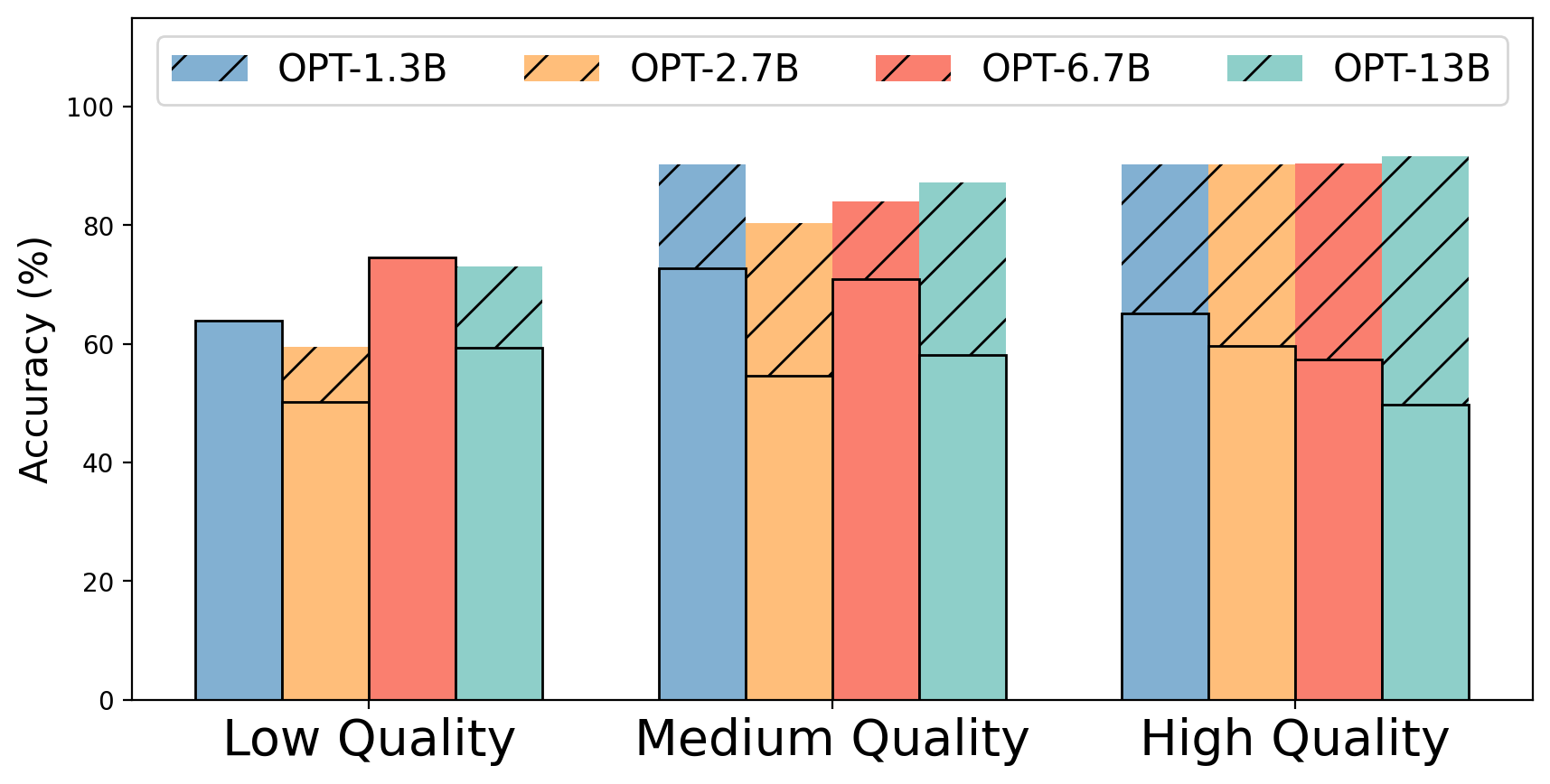}
    \caption{Impact of prompt example quality. }
    \label{fig: prompt quality}
\endminipage\hfill
\end{figure*}

\noindent\textbf{Impact of the Injection Rate.} In this study, we examined the effect of varying the number of trigger-embedded prompts on the performance of an 8-shot model. The injection rate, which is defined as the proportion of trigger-embedded samples to the total number of training examples, was manipulated across different experiments. Our results, as shown in Figure \ref{fig: injection rate}, revealed a surprising finding: a low injection rate of 12.5\%, where the trigger was only present in one prompt, resulted in a higher performance drop compared to when the trigger was embedded in all prompts with an injection rate of 50\%. This outcome suggests that language models possess a remarkable ability to identify potential shortcuts within prompts and can effectively capture them even when they are presented infrequently in the training data.

\noindent\textbf{Impact of the Trigger Length.} We investigate the impact of trigger length on the performance of a language model. Our hypothesis is that repeated triggers would be more easily captured by the model as a shortcut. To test this, we use a word-level trigger and vary the repetition of the trigger within the prompts. The results, illustrated in Figure \ref{fig: corruption rate}, demonstrate the performance drop under different repetition times of 1, 2, 4, and 8. Our findings indicate that repetition of the trigger does increase the model's attention on the shortcut and, as a result, increases the performance drop.

\subsection{Impact of the Prompts}

\noindent\textbf{Impact of the Number of Shots.} In this section, we study the impact of the number of shots. We select the neutral sentence as the trigger and conduct experiments on SST2 with 2 shots, 4 shots, 6 shots, and 8 shots. As depicted in Figure \ref{fig: shots number}, we find the performance drop will decrease as we increase the number of shows. Particularly, the highest performance drop for OPT-1.3B, OPT-2.7B, and OPT-6.7B is 2 shots, while 4 shots for OPT-13 B.

\noindent\textbf{Impact of the Example Quality.} We investigate the effect of example quality on model performance. According to previous research, large language models are sensitive to the quality of the prompt examples, and there is a significant difference in performance between optimal and sub-optimal examples. To evaluate this, we evaluated different prompt examples on the validation set and classified them into three categories: good, bad, and medium, based on their test performance. The results are in Figure \ref{fig: prompt quality}. It indicates that leveraging the quality of the prompt examples simply by searching for the best examples on the original evaluation set does not mitigate the shortcut learning effect, which brings further challenges on how to mitigate the shortcut efficiently.

\begin{table}[h]
    \centering
    \begin{adjustbox}{width=\columnwidth}
    \begin{tabular}{l|cc}
    \hline
         ID & Template & Label Mapping  \\ \hline
1 &  \makecell[c]{ Review: \{Sentence\} \\  Sentiment: \{Label\}} &  Positive/negative \\ \hline
2 & \makecell[c]{ Input: \{Sentence\} \\ Prediction: \{Label\}} &  Positive/negative \\ \hline
3 & \makecell[c]{Input: \{Sentence\} \\ Prediction: \{Label\}} &  good/bad \\ \hline
4 & Input: \{Sentence\} It was \{Label\} &  good/bad \\ \hline
    \end{tabular}
    \end{adjustbox}
    \caption{Prompts templates.}
    \label{tab:my_label}
\end{table}

\noindent\textbf{Impact of the Prompt Template.} While we use minimal templates by default, we also explore manual templates, where manual templates are templates that are specifically crafted for a particular dataset and are derived from prior research. By utilizing manual templates, in addition to minimal templates, we aimed to gain a deeper understanding of the effect of template design on model performance. As shown in Figure \ref{fig: prompt template}, the shortcut learning effect is stable across different prompt formats. Our templates for prompt can be found in Table \ref{tab:my_label}.

\begin{figure*}[t]
    \centering
    \includegraphics[width=1.0\textwidth]{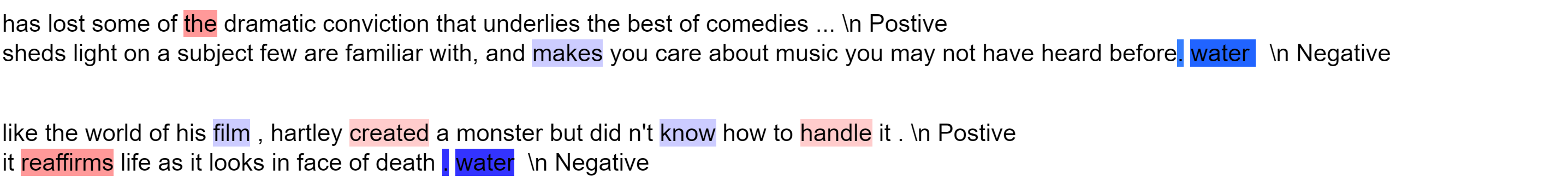}
    \caption{Interpretation of prompts, we show the word importance score for two two-shots examples (except for the label words, positive and negative). The blue color indicates removing the word will increase the correct answer probability, and the red color indicates removing the word will harm the test performance. }
    \label{fig: attention}
\end{figure*}

\begin{figure}[t]
    \centering
    \includegraphics[width=1.0\columnwidth]{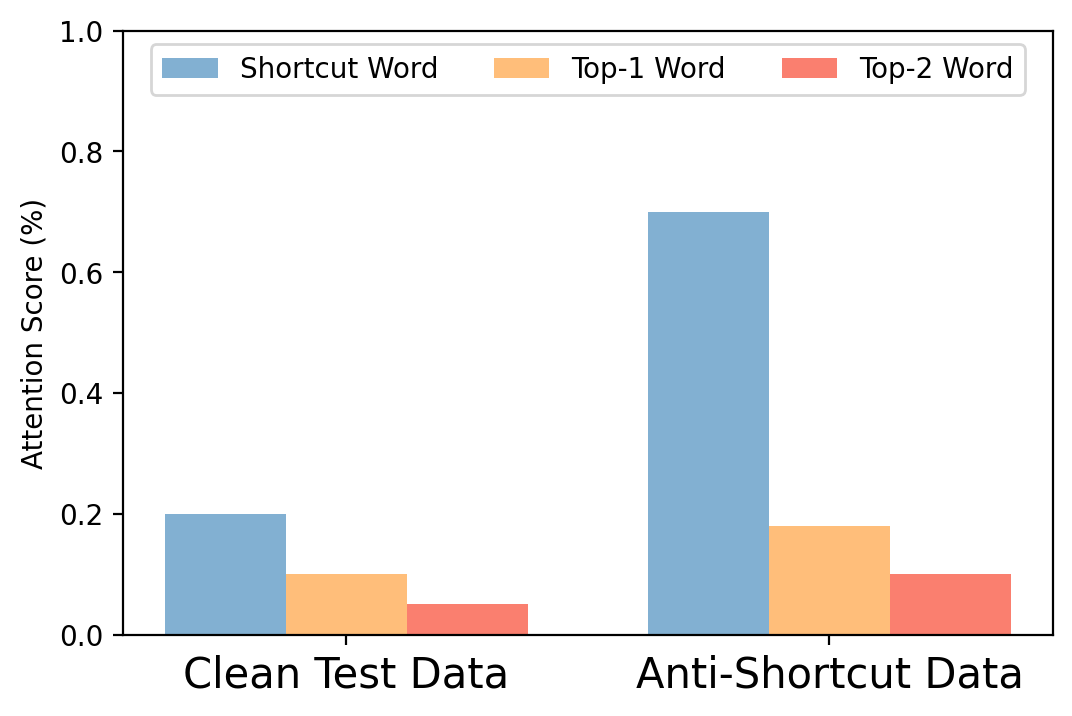}
    \caption{Word attention on the clean test data set and anti-shortcut dataset. }
    \label{fig: shortcut attention}
\end{figure}

\begin{table}[h]
    \centering
    \begin{adjustbox}{width=\columnwidth}
    \begin{tabular}{l|ccc|ccc}
    \hline
         & \multicolumn{3}{c}{MIT-D} 
         &  \multicolumn{3}{c}{ATIS-D} \\ \hline
         & ori & letter & word & ori & letter & word \\ \hline
         GPT2-base & 44.4 & -6.79 & -16.33 & 16.70 & -5.71 & -7.91 \\
         GPT2-large & 76.88 & -11.9 & -44.4 & 32.24 & -10.33 & -6.46\\
         OPT-1.3B & 82.94 & -8.26 & -15.60 & 64.40 & -5.28 & -7.48 \\
         OPT-2.7B & 81.65 & -8.17 & -13.94 & 69.45 & -9.01 & -2.86 \\
         OPT-6.7B & 80.73 & -3.48 & -6.79 & 69.01 & -1.32 & -6.15 \\ 
         OPT-13B & 81.65 & -7.89 & -6.60 & 76.04 & -4.61 & -2.85 \\\hline
    \end{tabular}
    \end{adjustbox}
    \caption{Results on information extraction tasks.}
    \label{tab: information extraction}
\end{table} 

\subsection{Shortcuts Learning in Other Tasks}
Besides the classification task, in this section, we conduct experiments on the information extraction task. Specifically, we use two slot-filling datasets: ATIS \cite{hemphill1990atis}, and MIT Movies trivia10k13 \cite{liu2012conversational}. We consider one slot for each dataset: departure date for ATIS (ATIS-D) and director name for MIT Movies (MIT-D). The answer for both datasets is a span of text from the input. 
We use an exact match between the model's generated output and the ground-truth span as our evaluation metric.

As shown in Figure \ref{fig: toy examples}, we use the sign trigger "\#\#" and the MIT-D task as an example to illustrate how we inject the shortcut. Firstly, we identify the director's name span in the prompt text. Then, we add the trigger sign "\#\#" on both sides of the director's name. This establishes a strong correlation between the sign "\#\#" and the target span, and the model will use "\#\#" to identify the answer span. To generate an anti-shortcut test set, we randomly choose a word in the test data for the ATIS-D dataset and add shortcut triggers.  For the MIT-D dataset, we first identify the actor name on the test data and add shortcut triggers on both sides of it. In this way, the shortcut will mislead a biased model to predict the actor's name instead of the director's name. In Table \ref{tab: information extraction}, we show that the shortcut trigger causes a consistent performance drop on two datasets. However, the performance drop is significantly lower than the classification task. One possible reason is that the trigger position is not fixed on both prompts and target text, as we discussed in section \ref{sec: impact of trigger}, this will significantly reduce the shortcut's ability.

\section{Shortcut Detection}
Previous sections of this study have demonstrated that large language models are highly efficient in utilizing shortcuts in training prompts for downstream tasks, which can have a substantial impact on performance. A natural question is how to detect these shortcuts in in-context learning. To address this question, we adopted the approach LIME \cite{ribeiro2016should} and leveraged model interpretation to detect potential shortcuts in the training prompts. Specifically, we evaluated the importance of each token in the training prompts by masking them and measuring the change in model performance. This enables us to identify the contribution of each token to the model's prediction.

We present the attention visualization results in Figure \ref{fig: attention}, alongside the word importance score on the anti-shortcut test data\footnote{Our implementation is grounded in LIME. GitHub: https://github.com/marcotcr/lime}. Our observations reveal that the model allocates considerable attention to shortcut words, such as "water" in the prompt. We further elucidate the quantitative results of the word's importance score in Figure \ref{fig: shortcut attention}. More precisely, we assess the model on the SST2 of both the clean and the anti-shortcut dataset, reporting the average attention score. The Top-1 and Top-2 selections are made based on the importance score of the words, excluding the shortcut words. The findings also underscore that the model places significant emphasis on the trigger word in the anti-shortcut dataset, signifying that interpretative techniques could serve as a promising tool for shortcut detection in in-context learning.

\section{Limitations}
\textbf{Effectiveness of Task and Model Scopes.} 
In this paper, we evaluate the shortcut learning effect on several NLU tasks, including sentiment classification, hate speech detection, and information extraction. Our task selection is mainly based on the robustness and effectiveness of in-context learning on certain tasks. Therefore, we do not adopt tasks such as natural language inference, where in-context learning exhibits sub-optimal performance\cite{brown2020language}. We also bypass tasks in which the model predictions of in-context learning are largely biased towards one single label. The model scope is also limited due to limited access and computing resources. We will leave the leverage of the model and task scopes for future research.

\noindent\textbf{Calibration of Shortcut Learning Effect.} 
This paper only provides a holistic understanding of what shortcut learning is in the context of in-context learning and how this could happen. Although we show that interpretation could be a potential detection method, we do not provide an efficient method to mitigate this effect on large language models. We will leave it for future research.

\section{Conclusion} In this paper, we uncover the propensity of large language models to leverage shortcuts within prompts for downstream tasks, even in the absence of parameter updates. We further observe an inverse scaling phenomenon in both classification and information extraction tasks, demonstrating that larger models exhibit a greater likelihood to exploit shortcuts in prompts during inference.

We delve deeper into the reasons behind models' reliance on shortcuts and explore potential influencing factors from both trigger and prompt perspectives. Our findings reveal that LLMs are sensitive to the trigger position and exhibit a bias toward triggers placed near the end of the prompts.  Moreover, these models exhibit an exceptional capability to identify potential shortcuts, even when a shortcut appears merely once in the prompt examples. Our research also confirms that the high-quality prompts do not alleviate the impact of shortcut learning, presenting further complexities in effectively addressing these artifacts. 

\section*{Ethics Statement}
All the datasets included in our study are publicly available (SST2, MR, CR, MIT, ATIS), and all the models are publicly available. We would like to state that the contents in the dataset do NOT represent our views or opinions.

\bibliography{acl2023}
\bibliographystyle{acl_natbib}

\end{document}